\documentclass[conference]{IEEEtran}
\usepackage{graphicx}

\usepackage{etoolbox}
\AtBeginEnvironment{quote}{\singlespacing\small}

\usepackage{setspace} 

\begin{document}
%
\title{Maximal Algorithmic Caliber and \\ Algorithmic Causal Network Inference: \\ General Principles of \\ Real-World General Intelligence?}

\author{\IEEEauthorblockN{Ben Goertzel}
\IEEEauthorblockA{SingularityNET Foundation\\
Suite 3-4, Tower 2, South Seas Centre\\
East Tsim Sha Tsui, Kowloon, Hong Kong\\
Email: ben@singularitynet.io}}


%


\maketitle

\begin{abstract}
Ideas and formalisms from far-from-equilibrium thermodynamics are ported to the context of stochastic computational processes, via following and extending Tadaki's algorithmic thermodynamics.   A Principle of Maximum Algorithmic Caliber is proposed, providing guidance as to what computational processes one should hypothesize if one is provided constraints to work within.   It is conjectured that, under suitable assumptions, computational processes obeying algorithmic Markov conditions will maximize algorithmic caliber.   It is proposed that in accordance with this, real-world cognitive systems may operate in substantial part by modeling their environments and choosing their actions to be (approximate and compactly represented) algorithmic Markov networks.   These ideas are suggested as potential early steps toward a general theory of the operation of pragmatic generally intelligent systems.
\end{abstract}


%
\IEEEpeerreviewmaketitle

\section{Introduction}

Given the successes of energy and entropy based formalisms in physics, it is natural to want to extend them into other domains like computation and cognition.   This is one natural path to follow in the quest to create general theories of real-world general intelligence, extending beyond theories such as Hutter's AIXI \cite{hutter2004universal} that model general intelligence elegantly but only under the assumption of unrealistically ample computational resources.

In this vein, the aim here is to sketch a workable approach to creating an energetic and entropic model of computational processes, and to indicate some of the potential implications of this model for cognitive systems. 

I will explain how one can articulate highly general principles of the dynamics of computational processes, that take a similar form to physics principles such as the stationary action principle (which often takes the form of "least action") and the Second Law of Thermodynamics (the principle of entropy non-decrease).   And I will explain how these principles may be used to understand the perceptual, cognitive and active aspects of intelligent systems.

In my recent work toward a general understanding of general intelligence, I have done things like formalizing Cognitive Synergy in terms of category theory \cite{DBLP:journals/corr/Goertzel17}, and articulating the Embodied Communication Prior in regard to which human-like agents attempt to be intelligent \cite{goertzel2009embodied}.   These ideas have elaborated key aspects of general intelligence, but have not given anything resembling a dynamical law of cognition.   On the other hand the ``cognitive equation'' I outlined in the 1990s \cite{Goertzel1994} is overly abstract, essentially formalizing the idea that a cognitive system iteratively recognizes patterns in its own structure and dynamics and then concretely instantiates these patterns within itself.  What has been lacking so far is an abstract yet reasonably precisely defined articulation of the dynamical laws of cognition.

Here I port several ideas from far-from-equilibrium thermodynamics  to the context of stochastic computational processes, via following and extending Tadaki's algorithmic thermodynamics \cite{tadaki2008statistical}.   I  propose a Principle of Maximum Algorithmic Caliber , extending Jaynes Maximum Caliber Principle, which provides  guidance as to what computational processes one should hypothesize if one is provided constraints to work within.  I then hypothesize that, under suitable assumptions, computational processes obeying algorithmic Markov conditions will maximize algorithmic caliber -- and that, in accordance with this, cognitive systems may operate by modeling their environments and choosing their actions to be (approximate and compactly represented) algorithmic Markov networks.

\section{Friston's Free Energy Principle}

Karl Friston's ``free energy principle'' represents one well-known effort in the direction of modeling cognition using physics-inspired principles.  It seems to me that Friston's ideas have some fundamental shortcomings -- but that eviewing these shortcomings has some value for understanding how to take a more workable approach.

I should clarify that the ideas presented here were not inspired by Friston's thinking to any degree, but more so by much older work in the  systems-theory literature -- e.g. Ilya Prigogine's Order out of Chaos \cite{prigogine2018order}, Eric Jantsch's The Self-Organizing Universe \cite{jantsch1980self}  and Hermann Haken's Synergetics \cite{haken2013principles}.    These authors represented a tradition within the complex-systems research community, of using far-from-equilibrium thermodynamics as a guide for thinking about life, the universe and everything.    Friston's ``free energy principle'' seems to have a somewhat similar conceptual orientation, but appears not to to incorporate the lessons of far-from-equilibrium thermodynamics very thoroughly, being based more on equilibrium thermodynamics concepts \cite{kirchhoff2018markov} \cite{parr2018anatomy}.   

A number of authors and researchers in the AI and neuroscience fields have expressed confusion regarding Friston's ideas, e.g., \cite{Alexander2018}  \cite{Schwarz2018}.   I share some of this confusion.   As the \cite{Schwarz2018} notes, regarding perception, Friston basically posits that neural and cognitive systems are engaged with trying to model the world they live in, and do so by looking for models with maximum probability conditioned on the data they've observed.   This is a useful but not adventurous perceptive, and one can formulate it in terms of trying to find models with  minimum KL-divergence to reality, which is one among many ways to describe Bayesian inference ... and which can be mathematically viewed as attempting to minimize a certain "free energy" function.

Friston then attempts to extend this principle to action via a notion of "active inference", and here things get even less clear   As \cite{kirchhoff2018markov} puts it,

\begin{quote}

Active inference is a cornerstone of the free energy principle. This principle states that for organisms to maintain their integrity they must minimize variational free energy.  Variational free energy bounds surprise because the former can be shown to be either greater than or equal to the latter. It follows that any organism that minimizes free energy thereby reduces surprise -- which is the same as saying that such an organism maximizes evidence for its own model, i.e. its own existence.

...

This interpretation means that changing internal states is equivalent to inferring the most probable, hidden causes of sensory signals in terms of expectations about states of the environment

...

[A] biological system must possess a generative model with temporal depth, which, in turn, implies that it can sample among different options and select the option that has the greatest (expected) evidence or least (expected) free energy. The options sampled from are intuitively probabilistic and future oriented. Hence, living systems are able to ?free? themselves from their proximal conditions by making inferences about probabilistic future states and acting so as to minimize the expected surprise (i.e. uncertainty) associated with those possible future states. This capacity connects biological qua homeostatic systems with autonomy, as the latter denotes an organism?s capacity to regulate its internal milieu in the face of an ever-changing environment. This means that if a system is autonomous it must also be adaptive, where adaptivity refers to an ability to operate differentially in certain circumstances. 

...

The key difference between mere and adaptive active inference rests upon selecting among different actions based upon deep (temporal) generative models that minimize the free energy expected under different courses of action.   This suggests that living systems can transcend their immediate present state and work towards occupying states with a free energy minimum.

\end{quote}

In this treatment, active inference is portrayed as a process by which cognitive systems take actions aimed at putting themselves in situations that will be minimally surprising, i.e. in which they will have the most accurate models of reality.    If taken literally this cannot be true, as it would predict that intelligent systems systematically seek simpler situations they can model better -- which is obviously not a full description of human motivation, for instance.   We do have a motivation to put ourselves in comprehensible, accurately model-able situations -- but we also have other motivations, such as the desire to perceive novelty and to challenge ourselves, which sometimes contradict our will to have a comprehensible environment.

One clear criticism of this analysis of active inference is that it's too much about states and not enough about paths.   To model far-from-equilibrium thermodynamics using energy-based formalisms, one needs to think about paths and path entropies and such, not just about things like ``work[ing] towards occupying states with a free energy minimum.''    Instead of thinking about ideas like ``selecting among different actions based upon deep (temporal) generative models that minimize the free energy expected under different courses of action.''  in terms of states with free energy  minimum, one needs to be thinking about action selection in terms of stationarity of action functions evaluated along multiple paths.
  
\section{Energetics for Far-From-Equilibrium Thermodynamics}

It seems clear that equilibrium thermodynamics isn't really what we want to use as a guide for cognitive information processing.  Fortunately, the recent thermodynamics literature contains some quite interesting results regarding path entropy in far-from-equilibrium thermodynamics.

Abaimov's paper  General formalism of non-equilibrium statistical mechanics, path approach \cite{abaimov2009general} and Raphael Chetrite and Hugo Touchette's paper Nonequilibrium Microcanonical and Canonical Ensembles and Their Equivalence  \cite{chetrite2013nonequilibrium} each tell part of the story.

David Rogers and Susan Rempe in \cite{rogers2011first} describe explicitly the far from equilibrium ``path free energy", but only for the case of processes with short memory, i.e. state at time $i+1$ depends on state $i$ but not earlier ones (which is often fine but not totally general).  

The following table from \cite{rogers2011first}  summarizes some key points concisely.

\begin{figure}
  \includegraphics[width=85mm]{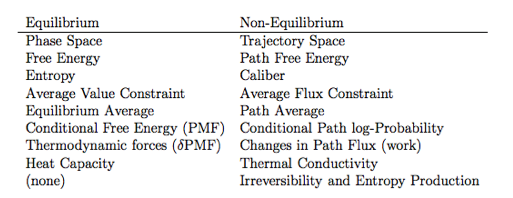}
  \caption{Correspondence between the single-time maximum entropy formulation of classical statistical mechanics, and the time dependent path maximum entropy formulation.}
  \label{fig:path-entropy}
\end{figure}

\noindent Conceptually, the crux is that we need to think not about the entropy of a state, but about the "caliber" of a path -- a normalization of the number of ways that path can be realized.   This then leads to the notion of the free energy of a certain path.     

It follows from this body of work that ideas like "free energy minimization" need to be re-thought dynamically rather than statically.   One needs to think about systems as following paths with differential probability based on the corresponding path free energies.    This is in line with the "Maximum Caliber principle"  \cite{dixit2018perspective} which is a generalization of the Maximum Entropy principle to dynamical systems (both first proposed in clear form by E.T. Jaynes, though Maximum Entropy has been more widely developed than Maximum Caliber so far).

Extending these notions further, Diego Gonzalez \cite{davis2015hamiltonian} outlines a Hamiltonian formalism that is equivalent to path entropy maximization, building on math from his earlier paper \cite{Gonzalez2014}.

\section{Action Selection and Active Inference}

Harking back to Friston for a moment, it follows that the dynamics of an intelligent system should be viewed, not as an attempt by an intelligent system to find a state with minimum free energy or surprisingness or any similar quantity, but rather as a process of a system evolving dynamically along paths chosen probabilistically to have stationary path free energy.   

But  of course, this would be just as true for an unintelligent system as for an intelligent system -- it's not a principle of intelligence but just a restatement of how physics works (in far from equilibrium cases; in equilibrium cases one can collapse paths to states).    

If we want to say something unique about intelligent systems in this context, we can look at the goals that an intelligent system is trying to achieve.   We may say that, along each potential path of the system's evolution, its various goals will be achieved to a certain degree.   The system then has can be viewed to have a certain utility distribution across paths -- some paths are more desirable to it than others.   A guiding principle of action selection would then be: To take an action A so that, conditioned on action A, the predicted probability distribution across paths is as close as possible to the distribution implied by the system's goals.

This principle of action selection can be formalized as KL-divergence minimization if one wishes, and in that sense it can be formulated as a "free energy minimization" principle.   But it's a "free energy" defined across ensembles of paths, not across states.

Relatedly, it's important to also understand that the desirability of a path to an intelligent system need not be expressible as the expected future utility at all moments of time along that path.   The desirability of a path may be some more holistic function of everything that happens along that path.    Considering only expected utility as a form of goal leads to various well known pathologies.

\section{Algorithmic Thermodynamics}

Next, how do we apply these same ideas beyond the realm of physics, to more general types of processes that change over time?

I am inspired by a general Whiteheadean notion of processes as fundamental things.   However, to keep things concrete, for now I'm going to provisionally assume that the``processes'' involved can be formulated as computer programs, in some standard Turing-equivalent framework, or maybe a quantum-computing framework.   I think the same ideas actually apply more broadly, but -- one step at a time...

Let us start with Kohtaro Tadaki's truly beautiful, simple, elegant paper titled A statistical mechanical interpretation of algorithmic information theory  \cite{tadaki2008statistical}.

Section 6 of Tadaki outlines a majorly aesthetic, obvious-in-hindsight parallel between algorithmic information theory and equilibrium thermodynamics.   There is seen to be a natural mapping between temperature in thermodynamics and compression ratio in algorithmic information theory.   A natural notion of ``algorithmic free energy''  is formulated, as a sort of weighted program-length over all possible computer programs (where the weights depend on the temperature).

The following table (drawn from Tadaki's presentation here \cite{Tadaki2008a}) summarizes the key  mappings in Tadaki's theory

 \begin{figure}
  \includegraphics[width=85mm]{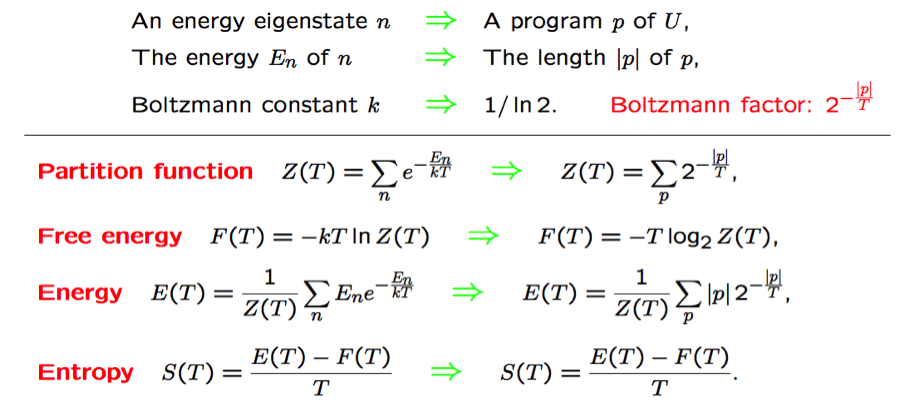}
  \caption{Correspondence between standard classical statistical mechanics and Tadaki's algorithmic statistical mechanics.}
  \label{fig:Tadaki}
\end{figure}

To ground the mappings he outlines, Tadaki gives a  simple statistical mechanical interpretation to algorithmic information theory.   He models an optimal computer as decoding equipment at the receiving end of a noiseless binary communication channel.   In this context, he regards programs for this computer as codewords (finite binary strings) and regards computation results (also finite binary strings) as decoded ?symbols.?    For simplicity he assumes that the infinite binary string sent through the channel -- constituting a series of codewords in a prefix-free code is generated by infinitely repeated tosses of a fair coin.   Based on this simple reductive model, Tadaki formulates computation-theoretic analogues to core constructs of traditional equilibrium thermodynamics.   

As a next step, it is clear one port the path-entropy based treatment of far-from-equilibrium thermodynamics as outlined above, to Tadaki's algorithmic-information context, by looking at sources emitting bits that are not independent of each other but rather have some probabilistic dependencies..

One thus obtains an ``algorithmic energy'' function that measures the energy of an algorithmic process over a period of time -- without assuming that it's a memoryless process like Tadaki does.   One also obtains a principle of Maximum Algorithmic Caliber, identical to the principle of Maximum Caliber but with algorithmic information in the place of Shannon entropy.   (To get this to work, one needs to make a few technical assumptions, e.g. one must assume that the knowledge one has of the dependencies among the bits produced by the process is given the form of expectations.)

This appears to lead to something Friston promises but doesn't actually deliver: A rigorous sense in which complex cognitive systems (modeled as algorithmic systems) are minimizing free energy.   The catch is that it?s an algorithmic path energy.

More precisely, relative to an observer $S$ who is observing a system $S_1$ in a  certain way (by tabulating conditional probabilities of ``how often some event of type $A$ occurs at time $T+s$, given some event of type $B$ occurred at time $T$``) , we may say the evolution of $S_1$ in $S$'s perspective obeys an energy minimization principle, where energy is defined algorithmic-informationally.

\subsection{Potential Quantum Extensions} 

Tadaki's treatment of algorithmic thermodynamics deals only with algorithms running on classical computers, but it appears his approach should be generalizable to the quantum case as well.     

In the quantum computing case, one is looking at series of qubits rather than bits, and instead of tabulating conditional probabilities one is tabulating amplitudes.   The maximum entropy principle is replaced with the stationary quantropy principle \cite{baez2015quantropy} and one still has the situation that: Relative to S who is observing S1 using some standard linear quantum observables, S1 may be said to evolve according to a stationary quantropy trajectory, where quantropy is here defined via generalizing the non-equilibrium generalization of Tadaki's algorithmic-informational entropy via replacing the real values with complex values.

\section{Algorithmic Causal Networks and Action Selection}

Ge et al \cite{ge2012markov} has shown that the maximal caliber is, under broad assumptions, achieved by Markov processes.    Specifically: When there are different possible dynamical trajectories in a time-homogeneous process, then the only type of process that maximizes the path entropy (given singlet statistics) is a sequence of  i.i.d. random variables, which is the simplest Markov process. If the data is in the form of sequentially pairwise statistics, then maximizing the caliber dictates that the process is Markovian with a uniform initial distribution; and if initial non-uniform dynamical distribution is known, or multiple trajectories are conditioned on an initial state, then the Markov process is still the only one that maximizes the caliber.  Similar results hold for processes that have history-dependence that extends over a history of a fixed finite maximum length.

It would seem that no one has yet derived similar results for the algorithmic caliber.   What sort of dynamic will yield maximal algorithmic caliber?   What is the proper algorithmic analogue of a Markov process?

Ganzing and Scholkopf \cite{janzing2010causal} define an ``algorithmic Markov condition'' that comprises  an analogue of a standard Bayesian causal network, but using conditional algorithmic information instead of conditional probability.   This is conceptually satisfying in the light of prior analyses of causality as a combination of temporal precedence and conditional probability, plus the presence of a simple explanation \cite{PLN}.      The algorithmic Markov condition embodies the ``simple explanation'' part of commonsense causality (where simplicity is represented by the  minimization of program length at the center of the definition of conditional algorithmic information), and then if one moves to the level of a statistical ensemble, one recovers conditional probability via the relationships between algorithmic information and Shannon entropy.

It is very natural to conjecture that {\bf maximum algorithmic caliber will be achieved via algorithmic processes whose state transition graphs obey a algorithmic Markov condition}.   I.e. if one needs to guess what program underlies certain observations, and one assumes the underlying program is time-homogeneous, then the most likely guess is going to be a program that agrees with these observations and whose state transition graph obeys an algorithmic Markov condition.    If an initial state is known then the conditional algorithmic informations used to define the algorithmic Markov condition can be additionally conditioned on the external state.   If the program is not time-homogeneous but depends on finite history, then this history-dependence can also be incorporated in the conditioning.

Supposing some variant of this conjecture is validated via detailed proof, there are potentially interesting implications for perception and action in cognitive systems, i.e. one is moved to suggest that

\begin{itemize}
\item A cognitive system should endeavor to model its environment as an algorithmic process whose state transition graph obeys an algorithmic Markov condition, which is consistent with its observations
\item A cognitive system should endeavor to select its actions according to a plan that obeys an algorithmic Markov condition, where the nodes in the algorithmic Markov network represent (context, action) pairs (so that a causal link between (a,c) and ($a_1$, $c_1$) in the network has the semantics `` action $a$ being taken in context $c$, implies action $a_1$ should be taken in context $c_1$''.    Given a collection of goals, the goal-driven activity of a cognitive system should be oriented to find a plan of this nature that has maximum simplicity, consistent with being maximally likely to achieve its goals.   
\item Cognition may be modeled in large part as perception of a system's internal knowledge base and action within this internal knowledge base; in this context the above two points apply to cognition as well
\item A key aspect of the reflective self-modeling  and social other-modeling of a cognitive system should be to construct an algorithmic Markov network model of itself and of other agents.   Constructing an algorithmic Markov model of its own prior actions and then executing new actions according to this model, is a meaningful form of not-necessarily-goal driven activity in a complex, intelligent system.
\end{itemize}

A few caveats, however, are also critical here:

\begin{itemize}
\item Algorithmic information being uncomputable in general, real cognitive systems will need to more or less roughly approximate principles like these in practice.    Principles like Cognitive Synergy \cite{DBLP:journals/corr/Goertzel17} may be considered partially as strategies for approximating the learning of causal algorithmic networks under limited resources in the context of certain sorts of environments and goals.
\item The above suggestions are statements about  mathematical representability rather than about efficient pragmatic knowledge representation.   It may be that the algorithmic Markov networks required for a certain cognitive system are best represented in some other more compact form, rather than explicitly as algorithmic Markov networks.   For instance a hypergraph like the OpenCog Atomspace \cite{EGI1} \cite{EGI2} can be used as a compacted way of doing storage and retrieval for a large number of complexly overlapping algorithmic Markov networks dealing with perceptual data, actions and action plans, or more abstract cognitions.
\item The mapping between  maximum algorithmic caliber and algorithmic Markov conditions has certain prerequisite conditions, like time-homogeneity or limited-scope history-dependence.  Furthermore the heuristic of maximum algorithmic caliber is rough-and-ready and valuable for cognitive systems, but isn't the answer to every problem an intelligent system faces.   Sometimes the best way to achieve system goals will involve learning patterns and programs other than algorithmic causal networks.
\end{itemize}

\noindent What is proposed here is not a universal simplified solution to the task of general intelligence under limited resources, but rather an abstract formulation of a cognitive principle which, via various pragmatic approximations, seems capable to explain a large percentage of what a generally intelligent cognitive system needs to do in practical settings.   

Specifically if we think about the Embodied Communication Prior \cite{goertzel2009embodied} mentioned above, i.e. the task of being generally intelligent in the context of controlling social communicative agents that share physical environments with other social communicative agents, then it is hypothesized that the vast majority of perceptions and actions and a high percentage of cognitions carried out by systems that are generally intelligent in this setting can be explained as approximations to the learning of causal algorithmic networks as outlined here.

\section{Conclusion}

Connecting together a number of threads from recent research in physics and computer science, we have articulated some potential general principles governing the activity of generally intelligent systems.   Considerable mathematical work will need to be done to transform the ideas and conjectures outlined here into rigorous definitions, theorems and proofs.   However, we believe this is a valuable and worthwhile direction for research; the key idea being to take key concepts from far-from-equilibrium thermodynamics and port them to algorithmic information theory, and using this methodology to articulate laws and principles for (computationally-modeled) cognitive activity, inspired by laws and principles of physics together with concepts of cognitive modeling.   Proposals such as algorithmic Markov networks to model perception, action and self-modeling have been presented here as initial examples of cognitive theorization along these lines.

\bibliographystyle{IEEEtran}
\bibliography{bbm}
%



\end{document}